\documentclass[letterpaper]{article} 
\usepackage{aaai2027} 
\usepackage[hyphens]{url} 
\usepackage{graphicx} 
\usepackage{natbib} 
\usepackage{caption} 
\usepackage{amsmath,amssymb,bm}
\usepackage{booktabs}
\usepackage{multirow}
\usepackage{xcolor}
\usepackage{colortbl}
\usepackage{placeins}
\usepackage{cuted}


\newcommand{\safeincludegraphics}[2][]{%
  \IfFileExists{#2}{%
    \includegraphics[#1]{#2}%
  }{%
    \IfFileExists{#2.pdf}{%
      \includegraphics[#1]{#2.pdf}%
    }{%
      \IfFileExists{#2.png}{%
        \includegraphics[#1]{#2.png}%
      }{%
        \IfFileExists{#2.jpg}{%
          \includegraphics[#1]{#2.jpg}%
        }{%
          \IfFileExists{#2.jpeg}{%
            \includegraphics[#1]{#2.jpeg}%
          }{%
            \fbox{%
              \rule{0pt}{0.055\textwidth}%
              \rule{0.055\textwidth}{0pt}%
            }%
          }%
        }%
      }%
    }%
  }%
}

\title{Text-Guided Visual Dependency Graph Learning with Cross-Modal Attention Priors}

\author{
Fei Wang\textsuperscript{\rm 1},
Yutong Zhang\textsuperscript{\rm 2},
Yang Ye\textsuperscript{\rm 3},\\
Jinxian Chen\textsuperscript{\rm 3},
Wang Wenshuai\textsuperscript{\rm 3},
Xiong Wang\textsuperscript{\rm 4}
}
\affiliations{
\textsuperscript{\rm 1}Stony Brook University, USA,\\
\textsuperscript{\rm 2}Sichuan University, China,\\
\textsuperscript{\rm 3}Universiti Sains Malaysia, Malaysia\\
\textsuperscript{\rm 4}University of Science and Technology of China, China
}

\begin{document}
\maketitle

\begin{abstract}
Estimating interpretable conditional-dependence structures from multimodal visual--linguistic features remains largely unexplored.
We propose CM-GLasso (Cross-Modal Graphical Lasso), a framework that bridges vision--language representation learning and sparse Gaussian Graphical Models.
CM-GLasso introduces three key components:
(i)~a \textit{text visualization} strategy that renders class-attribute descriptions as images and processes them through the same SigLIP~2 vision encoder as natural images, yielding prototype-indexed patch-level attention footprints in a shared feature coordinate system;
(ii)~a \textit{cross-attention distillation} mechanism that condenses high-dimensional patches into a small set of semantic graph nodes, whose attention-footprint similarities yield cross-modal structural priors for non-uniform $\ell_1$ penalization;
(iii)~a \textit{joint ADMM formulation} that estimates shared and class-specific precision components within a single convex objective, avoiding the need to first estimate and then decompose separate class-wise graphs.
The learned sparse graph topologies directly support a parameter-free, precision-based classification rule and a lightweight topology-aware segmentation head.
Extensive experiments on eight benchmarks demonstrate that CM-GLasso achieves competitive or superior performance compared with strong feature-based and task-specific baselines. Under the matched controlled protocol, it attains the highest average classification accuracy (91.97\%) and the highest segmentation mIoU among the controlled baselines on VOC (74.75\%) and ADE20K (64.01\%), while also yielding explicit sparse conditional-dependence graphs with common--specific decomposition.
\end{abstract}

\section{Introduction}
\label{sec:intro}

Sparse conditional-dependence graphs provide an interpretable alternative to opaque feature interactions, but extending Graphical Lasso (GLasso)~\cite{friedman2008sparse} to multimodal vision raises three difficulties. First, class-conditional covariance estimates are unstable when the sample size is small relative to the graph dimension ($n_c<p$ for some classes). Second, independently encoded modalities do not provide dimensionally aligned structural priors. Third, separate class-wise graphs fail to distinguish invariant structure from class-specific deviations. Prior-weighted estimators such as Tailored GLasso~\cite{lingjarde2021tailored} address auxiliary information in unimodal settings, but not vision--language attention or common--specific visual graphs.

We propose \textbf{CM-GLasso} (Figure~\ref{fig:pipeline}). Text descriptions are rendered as images and, together with natural images, processed by the same SigLIP~2 vision encoder~\cite{tschannen2025siglip2}. Shared semantic prototypes distill patch tokens into $p$ graph nodes; overlap between prototype attention footprints produces a class-conditioned $p\times p$ prior that weights the sparsity penalty. A single convex objective then estimates a shared precision component and class-specific deviations. Its prior strength is selected by eBIC: $k^*=0$ yields a uniformly penalized joint GLasso--CSSL model rather than a single-graph GLasso. ADMM~\cite{boyd2011distributed} solves the resulting program, and the learned graphs directly support generative classification and topology-aware segmentation.

Our contributions are:
\vspace{-1mm}
\begin{itemize}
    \item a shared-encoder text-visualization and cross-attention distillation scheme that constructs dimensionally aligned multimodal structural priors;
    \item a prior-weighted common--specific precision-matrix objective with eBIC-controlled prior use and joint ADMM estimation; and
    \item graph-based classification and segmentation heads evaluated on eight benchmarks, providing competitive performance together with explicit sparse topologies.
\end{itemize}
\vspace{-1mm}

\section{Related Work}

\textbf{Sparse Precision Matrix Estimation and CSSL.}
GLasso~\cite{friedman2008sparse}, non-uniform penalties~\cite{ambroise2009inferring}, and eBIC-guided selection~\cite{foygel2010extended, lingjarde2021tailored} improve sparse precision estimation. CSSL~\cite{hara2013learning} and joint graphical models~\cite{danaher2014joint} separate shared and condition-specific structure, while sparse precision matrices have also modeled visual label interactions~\cite{souly2016scene}. CM-GLasso adds vision--language attention priors and graph-structured downstream inference to this common--specific formulation.

\textbf{Vision-Language Representation and Adaptation.}
CLIP~\cite{radford2021learning} and SigLIP~2~\cite{tschannen2025siglip2} align visual and linguistic representations; CoOp~\cite{zhou2022coop}, TIP-Adapter~\cite{zhang2022tipadapter}, and DINOv2 probes~\cite{oquab2024dinov2} are strong adaptation baselines. Prototype learning~\cite{lu2024palm, wei2023onpro} compresses features into representative vectors. Unlike these methods and general multimodal architectures~\cite{steiner2024paligemma2,wu2025ja}, CM-GLasso uses shared prototype-indexed attention geometry to define nodes and transfer graph priors.

\textbf{Visual Text Compression (VTC).}
VTC renders text for efficient long-context decoding~\cite{wei2025textorpixels,cheng2025glyph}. CM-GLasso instead renders short, offline class descriptions solely to obtain patch-level attention footprints; it performs neither token compression nor online LLM decoding.

\textbf{Deep Structural Learning for Downstream Vision Tasks.}
Segmentation models~\cite{jain2023oneformer,wang2023internimage,xu2022groupvit,lin2023clipes,jia2024polypmixnet} and prompt-tuned classifiers~\cite{shang2025provpt,ren2025davpt} capture contextual interactions but do not expose sparse conditional-dependence graphs. CM-GLasso supplies this explicit common--specific topology as an interpretable inductive bias.

\section{Methodology}
\label{sec:method}

\begin{figure*}[t]
    \centering
    \includegraphics[width=\textwidth]{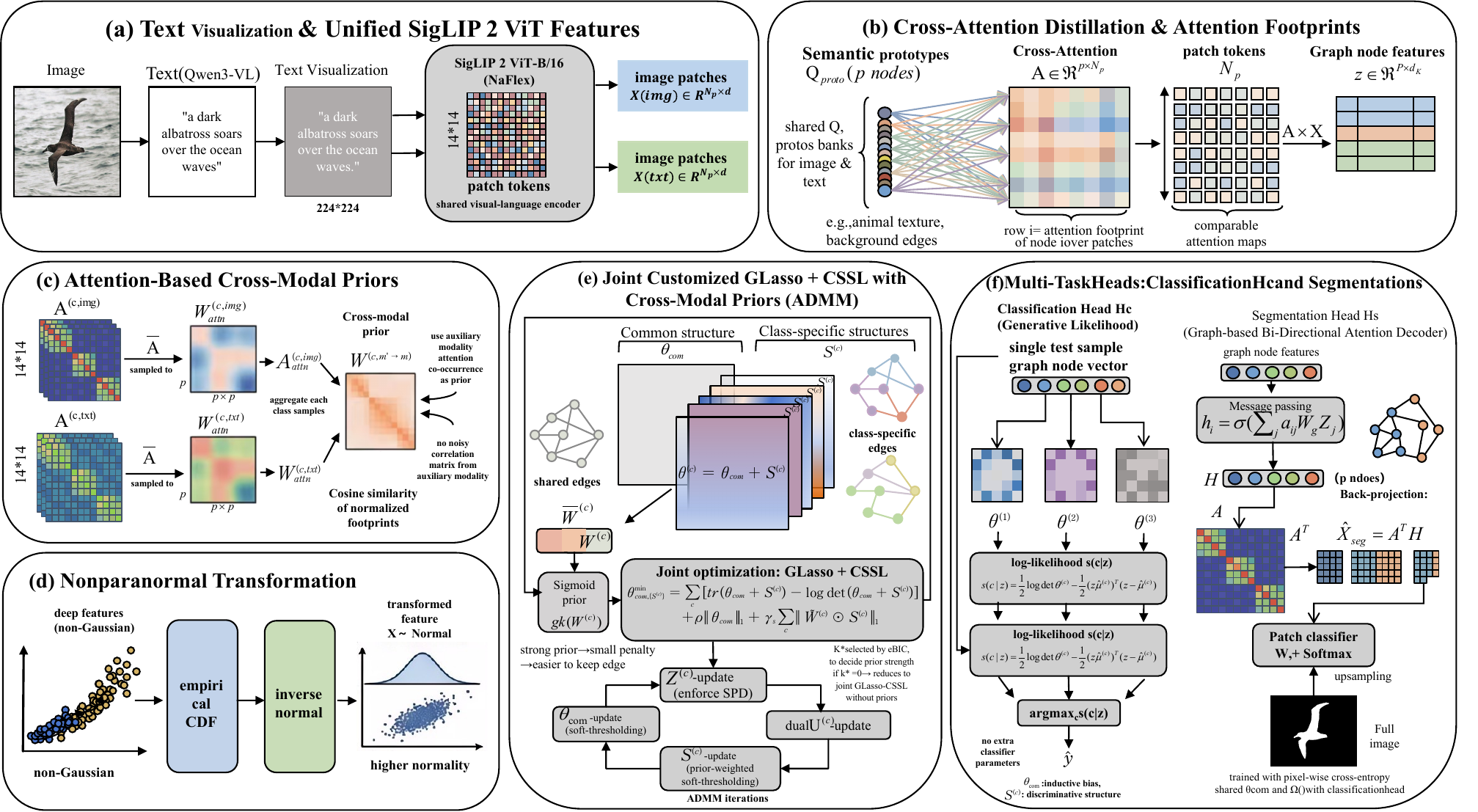}
    \caption{Overview of the CM-GLasso pipeline. (a) Text visualization and unified SigLIP~2 feature extraction. (b) Cross-attention distillation condenses patch tokens into semantic nodes. (c) Prototype attention footprints produce spatially aware cross-modal priors. (d--e) A rank-based nonparanormal transformation improves marginal Gaussianity, after which joint ADMM estimation separates the shared $\boldsymbol{\Theta}_{\mathrm{com}}$ and class-specific $\boldsymbol{S}^{(c)}$ components. (f) The learned structures guide generative classification and topology-aware segmentation.}
    \label{fig:pipeline}
\end{figure*}

Figure~\ref{fig:pipeline} summarizes three stages: shared-encoder cross-modal prior construction, nonparanormal common--specific graph estimation, and graph-structured classification/segmentation.

\subsection{Problem Formulation}

For each class $c$, we estimate the image-domain precision matrix $\boldsymbol{\Theta}^{(c)}=(\boldsymbol{\Sigma}^{(c)})^{-1}$ from $n_c$ training observations and decompose it as $\boldsymbol{\Theta}^{(c)}=\boldsymbol{\Theta}_{\mathrm{com}}+\boldsymbol{S}^{(c)}$~\cite{hara2013learning,danaher2014joint}. Rendered text does not define a separate target precision matrix; it supplies only the cross-modal penalty prior. An off-diagonal $\theta_{ij}\neq0$ denotes conditional dependence, with partial correlation $-\theta_{ij}/\sqrt{\theta_{ii}\theta_{jj}}$. Its sign represents statistical association rather than causation; we use sign-separated edges only as a message-passing inductive bias.

\subsection{Unified Representation and Graph Nodes}

Offline class descriptions are rendered as $224\times224$ text images and processed with natural images by the same SigLIP~2 ViT-B/16 NaFlex vision tower~\cite{tschannen2025siglip2,zhai2023sigmoid}. Using pre-MAP patch states gives $\mathbf{X}^{(m)}\in\mathbb{R}^{N_p\times d}$ for $m\in\{\mathrm{img},\mathrm{txt}\}$, with $N_p=196$ and $d=768$. Thus both modalities share encoder weights and prototype indices; no online LLM decoding is used.

Class descriptions are generated once, before training, and rendered with
adaptive font sizing as black text on a white canvas. This differs from visual
text compression: the goal is not to reduce language-token count, but to
obtain spatially resolved patch responses from the same pathway used for
natural images. Sharing the vision tower removes an additional cross-encoder
alignment module and makes prototype index $i$ comparable across modalities.
SigLIP~2 uses a MAP pooling head rather than a dedicated \texttt{[CLS]} token;
we therefore take the patch states immediately before MAP pooling. The NaFlex
checkpoint preserves aspect-ratio information while enforcing a maximum of
196 patches, providing a consistent interface for rendered descriptions and
resized natural images.

\subsubsection{Cross-Attention Distillation: From Patches to Graph Nodes}
\label{sec:cross_attn_distill}

Learnable prototypes $\mathbf{Q}_{\mathrm{proto}}\in\mathbb{R}^{p\times d}$ distill patches into graph nodes:
\begin{equation}
\begin{aligned}
\mathbf{A}&=\operatorname{softmax}\!\left(\frac{(\mathbf{Q}_{p}\mathbf{W}_Q)(\mathbf{X}\mathbf{W}_K)^\top}{\sqrt{d_k}}\right),\\
\mathbf{Z}&=\mathbf{A}(\mathbf{X}\mathbf{W}_V),\qquad
\mathbf{z}_n=\mathbf{Z}_n\mathbf{w}_{\mathrm{out}}\in\mathbb{R}^{p}.
\end{aligned}
\label{eq:cross_attn}
\end{equation}
Rows of $\mathbf{A}\in\mathbb{R}^{p\times N_p}$ are prototype attention footprints. Shared prototypes align node indices across modalities, while $\mathbf{z}_n$ provides the observation used by GLasso.

\subsubsection{Construction of the Prior Matrix}
\label{sec:prior_from_attn}

For class $c$, we aggregate and row-normalize attention, then use auxiliary-modality cosine overlap as the target-graph prior:
\begin{equation}
\begin{aligned}
\bar{\mathbf{A}}_{i,:}^{(c,m)}&=
\frac{\mathbf{A}_{\mathrm{agg},i,:}^{(c,m)}}{\|\mathbf{A}_{\mathrm{agg},i,:}^{(c,m)}\|_2},\\
\mathbf{W}^{(c,m'\to m)}&=
\bar{\mathbf{A}}^{(c,m')}(\bar{\mathbf{A}}^{(c,m')})^\top.
\end{aligned}
\label{eq:cross_prior}
\end{equation}
This heuristic prior does not fix graph support; it only weights sparsity, leaving edges to the target likelihood and common--specific decomposition.

\subsection{Nonparanormal Transformation}
\label{sec:nonparanormal}

Transformer nodes are non-Gaussian (only $\sim23\%$ pass Shapiro--Wilk at $\alpha=.05$). Following~\cite{liu2009nonparanormal}, each class-conditional marginal is rank-Gaussianized:
\begin{equation}
\begin{aligned}
\tilde z_{nj}^{(c)}&=\Phi^{-1}\!\left(\hat F_j^{(c)}(z_{nj})\right),\\
\hat F_j^{(c)}(z_{nj})&=
\frac{\operatorname{rank}^{(c)}(z_{nj})-.5}{n_c}.
\end{aligned}
\label{eq:nonparanormal}
\end{equation}
Training CDFs, clipped with the standard Winsorization bound, transform each test sample for every candidate class. Because these class-specific transforms differ, Eq.~\eqref{eq:cls_head} is an approximate transformed-space discriminant that omits the Jacobian. Joint Gaussianity remains a modeling assumption: the held-out marginal pass rate rises to $\sim88\%$, but residual non-Gaussianity remains.

We compute the centered class covariance
\begin{equation}
\hat{\boldsymbol{\Sigma}}^{(c)}=
\frac{1}{n_c-1}\sum_{n:y_n=c}
(\tilde{\mathbf z}_n-\hat{\boldsymbol\mu}^{(c)})
(\tilde{\mathbf z}_n-\hat{\boldsymbol\mu}^{(c)})^\top.
\label{eq:empirical_cov}
\end{equation}

\textbf{Test-time transformation.}
For each candidate class $c$, a test observation is transformed with the CDF
fitted only on that class's training samples:
\begin{equation}
\begin{aligned}
\tilde z_{*j}^{(c)}&=
\Phi^{-1}\!\left[
\operatorname{clip}\!\left(
\hat F_j^{(c)}(z_{*j}),\delta_c,1-\delta_c
\right)\right],\\
\delta_c&=\frac{1}{4n_c^{1/4}\sqrt{\pi\log n_c}}.
\end{aligned}
\label{eq:test_npn}
\end{equation}
The CDF is never re-estimated from test data. Class-dependent transforms mean
that candidate scores are evaluated in different Gaussianized coordinates;
consequently, the classifier is a transformed-space discriminant rather than
an exact likelihood in the original node space. The transformation guarantees
neither a jointly Gaussian distribution nor a correct nonparanormal model.
We therefore report the held-out Shapiro--Wilk pass rate as a diagnostic, not
as proof of the modeling assumption. Dimensions that remain non-Gaussian can
arise from multimodal node responses or small class sample sizes.

\raggedbottom
\subsection{Unified Optimization of Tailored GLasso and CSSL}
\label{sec:joint_opt}

Images are the estimation target and rendered text supplies the prior. Writing $\boldsymbol{\Theta}^{(c)}=\boldsymbol{\Theta}_{\mathrm{com}}+\boldsymbol{S}^{(c)}$, we jointly estimate shared structure and class-specific deviations rather than decomposing fixed class-wise graphs~\cite{danaher2014joint,hara2013learning}:
\begin{equation}
\label{eq:joint}
\begin{aligned}
\min_{\boldsymbol{\Theta}_{\text{com}},\{\boldsymbol{S}^{(c)}\}}
\quad
&
\sum_{c=1}^{C}
\Big[
\operatorname{tr}\left(
\hat{\boldsymbol{\Sigma}}^{(c)}
(\boldsymbol{\Theta}_{\text{com}}+\boldsymbol{S}^{(c)})
\right)
\\
&\qquad
-\log\det(
\boldsymbol{\Theta}_{\text{com}}+\boldsymbol{S}^{(c)}
)
\Big]
\\
&+
\rho
\|\boldsymbol{\Theta}_{\text{com}}\|_{1,\text{off}}
\\
&+
\gamma_s
\sum_{c=1}^{C}
\|
\tilde{\mathbf{W}}^{(c)}
\odot
\boldsymbol{S}^{(c)}
\|_{1,\text{off}}
\\
\text{s.t.}\quad
&
\boldsymbol{\Theta}_{\text{com}}
+
\boldsymbol{S}^{(c)}
\succ0,
\quad\forall c.
\end{aligned}
\end{equation}

The eBIC-selected sigmoid converts attention overlap into penalty weights:
\begin{equation}
\label{eq:adaptive_weight}
\tilde{w}_{ij}^{(c)}
=
1-
\frac{1}{
1+\exp\left(
-k^*(W_{ij}^{(c,m'\to m)}-0.5)
\right)
}.
\end{equation}

\subsubsection{Efficient Optimization via ADMM}
\label{sec:admm}

ADMM~\cite{boyd2011distributed} introduces $\mathbf Z^{(c)}=\boldsymbol\Theta_{\mathrm{com}}+\boldsymbol S^{(c)}$ and unscaled dual variables $\mathbf U^{(c)}$. The $\mathbf Z$ step is the standard log-determinant proximal eigensolution; the remaining steps are off-diagonal soft-thresholding:
\begin{equation}
\begin{aligned}
\boldsymbol\Theta_{\mathrm{com}}&\leftarrow
\mathcal S^{\mathrm{off}}_{\rho/(C\mu)}\!\left[
\frac1C\sum_c(\mathbf Z^{(c)}-\boldsymbol S^{(c)}+\mu^{-1}\mathbf U^{(c)})\right],\\
\boldsymbol S^{(c)}&\leftarrow
\mathcal S^{\mathrm{off}}_{\gamma_s\tilde{\mathbf W}^{(c)}/\mu}
(\mathbf Z^{(c)}-\boldsymbol\Theta_{\mathrm{com}}+\mu^{-1}\mathbf U^{(c)}),\\
\mathbf U^{(c)}&\leftarrow\mathbf U^{(c)}+\mu(\mathbf Z^{(c)}-\boldsymbol\Theta_{\mathrm{com}}-\boldsymbol S^{(c)}).
\end{aligned}
\label{eq:admm_updates}
\end{equation}

For completeness, define
$\mathbf G^{(c)}=\boldsymbol\Theta_{\mathrm{com}}+\boldsymbol S^{(c)}-\mu^{-1}\mathbf U^{(c)}$.
The positive-definite block solves
\begin{equation}
\begin{aligned}
\mathbf Z^{(c)}\leftarrow
\arg\min_{\mathbf Z\succ0}\;&
\left\{
\operatorname{tr}(\hat{\boldsymbol\Sigma}^{(c)}\mathbf Z)
-\log\det\mathbf Z
\right.\\[-1mm]
&\left.\qquad+\frac{\mu}{2}\|\mathbf Z-\mathbf G^{(c)}\|_F^2
\right\}.
\end{aligned}
\label{eq:z_update}
\end{equation}
If $\mathbf Q\boldsymbol\Lambda\mathbf Q^\top$ is the eigendecomposition of
$\mathbf G^{(c)}-\mu^{-1}\hat{\boldsymbol\Sigma}^{(c)}$, then
\begin{equation}
\mathbf Z^{(c)}=\mathbf Q\widetilde{\boldsymbol\Lambda}\mathbf Q^\top,
\qquad
\tilde\lambda_i=\frac{\lambda_i+\sqrt{\lambda_i^2+4/\mu}}{2}.
\label{eq:z_eigen}
\end{equation}
This update enforces positive definiteness at every iteration, so we use
$\hat{\boldsymbol\Theta}^{(c)}=\mathbf Z^{(c)}$ for downstream inference.
Iterations stop when both primal and dual residuals fall below their absolute
and relative tolerances, or after 200 iterations. We search $k$ by eBIC on
training data; when $k^*=0$, all weights equal $1/2$, giving prior-free joint
GLasso--CSSL. The resulting optimization has three primal blocks, so the
general convergence theorem for two-block ADMM does not directly apply,
although all reported runs converged stably.

\subsubsection{Model Selection and Numerical Stability}

For each prior direction, $k$ is selected from the prescribed grid by the
extended Bayesian information criterion~\cite{foygel2010extended}:
\begin{equation}
\operatorname{eBIC}(k)=
-2\ell(\hat{\boldsymbol\Theta}_k)
+|E_k|\log n
+4\gamma |E_k|\log p,
\label{eq:ebic}
\end{equation}
where $E_k$ is the estimated off-diagonal support and the same $\gamma$ is
used for every candidate. Selection is performed using training observations;
validation data are used only for the outer choices of $\rho$ and $\gamma_s$.
The frequent selection of $k^*=0$ for self-priors (Table~\ref{tab:all_ablations})
shows that the procedure can reject auxiliary structure rather than forcing it
into the estimate.

When $n_c<p$, the empirical covariance is singular. We apply a small diagonal
loading before optimization, symmetrize every matrix update, and preserve the
unpenalized diagonal during soft-thresholding. At iteration $t$, convergence is
monitored with
\begin{equation}
\begin{aligned}
r_t&=\max_c\|\mathbf Z_t^{(c)}-\boldsymbol\Theta_{\mathrm{com},t}
-\boldsymbol S_t^{(c)}\|_F,\\
s_t&=\mu\max_c\|\boldsymbol\Theta_t^{(c)}-
\boldsymbol\Theta_{t-1}^{(c)}\|_F.
\end{aligned}
\label{eq:admm_residuals}
\end{equation}
The eigensolution in Eq.~\eqref{eq:z_eigen} is also used as the final
precision matrix, avoiding a possibly indefinite reconstruction from separately
thresholded common and specific blocks.

\subsection{Multi-task Heads: From Graph Structures to Downstream Predictions}
\label{sec:downstream}

The learned graphs drive both tasks.

\subsubsection{Classification Head $\mathcal{H}_C$: Generative Discrimination}
\label{sec:cls_head}

For downstream inference, we use the positive-definite ADMM variable $\hat{\boldsymbol\Theta}^{(c)}=\mathbf Z^{(c)}$. The transformed-space classification score is
\begin{equation}
\label{eq:cls_head}
\begin{aligned}
s_c(\mathbf{z}_*)
={}&
\frac{1}{2}
\log\det\hat{\boldsymbol{\Theta}}^{(c)}
\\
&-
\frac{1}{2}
(\tilde{\mathbf{z}}_*^{(c)}-\hat{\boldsymbol{\mu}}^{(c)})^{\top}
\hat{\boldsymbol{\Theta}}^{(c)}
(\tilde{\mathbf{z}}_*^{(c)}-\hat{\boldsymbol{\mu}}^{(c)})
+
\log\pi_c.
\end{aligned}
\end{equation}

The predicted label is $\hat y=\arg\max_c s_c(\mathbf z_*)$.

\subsubsection{Segmentation Head $\mathcal{H}_S$: Graph-Structured Attention Decoding}
\label{sec:seg_head}

For each candidate class, positive and negative entries of the final precision matrix $\hat{\boldsymbol{\Theta}}^{(c)}=\mathbf Z^{(c)}$ use separate transformations:
\begin{equation}
\begin{aligned}
\mathbf{h}_i^{(c)}
&=\sigma\left(
\sum_{j:\hat\theta_{ij}^{(c)}>0}
\alpha_{ij}^{+}\mathbf{W}_{\text{pos}}\mathbf{Z}_{n,j}
\right.\\[-1mm]
&\qquad\left. +
\sum_{j:\hat\theta_{ij}^{(c)}<0}
\alpha_{ij}^{-}\mathbf{W}_{\text{neg}}\mathbf{Z}_{n,j}
\right).
\end{aligned}
\label{eq:graph_mp_signed}
\end{equation}

The normalized coefficients are derived directly from precision magnitudes:
\begin{equation}
\alpha_{ij}^{+}=
\frac{[\hat\theta_{ij}^{(c)}]_+}{\sum_{q\neq i}[\hat\theta_{iq}^{(c)}]_++\varepsilon},
\qquad
\alpha_{ij}^{-}=
\frac{[-\hat\theta_{ij}^{(c)}]_+}{\sum_{q\neq i}[-\hat\theta_{iq}^{(c)}]_++\varepsilon}.
\label{eq:signed_norm}
\end{equation}
This separates positively and negatively associated pathways without treating
the sign as a causal relation. A residual self-node pathway retains local
evidence when a row is sparse.

Attention back-projects nodes and produces patch logits:
\begin{equation}
\hat{\mathbf X}_n^{(c)}=\mathbf A^\top\mathbf H_n^{(c)},
\qquad
\boldsymbol\ell_{n,:,c}=\hat{\mathbf X}_n^{(c)}\mathbf w_s^{(c)}.
\end{equation}
Patch logits are reshaped to the encoder grid and bilinearly upsampled. The
decoder is trained with pixel-wise cross-entropy while the estimated graph is
fixed; thus no segmentation label is used to refit the precision matrices.

\subsection{Decoupled Training Pipeline}

Training proceeds in four phases. First, a temporary supervised head trains
the semantic prototypes and projection matrices, ensuring that distilled nodes
retain task-relevant information. Second, the node extractor is frozen and
training-only observations are collected to estimate
$\boldsymbol\Theta_{\mathrm{com}}$ and $\{\boldsymbol S^{(c)}\}$ offline.
Third, these graphs remain fixed while the segmentation back-projection and
patch classifier are fitted. Finally, inference freezes both neural parameters
and graph structures. This separation prevents validation or test observations
from entering covariance, CDF, prior-selection, or graph-estimation steps, but
it does not yield a globally optimal joint neural--graph solution.

\flushbottom
\section{Experiments}
\label{sec:experiments}

\subsection{Experimental Setup}

\textbf{Datasets:}
We evaluate on eight benchmarks: CIFAR-10/100~\cite{krizhevsky2009learning}, CUB-200-2011~\cite{wah2011cub}, and Caltech-256~\cite{griffin2007caltech256} for classification; PASCAL VOC 2012~\cite{everingham2012pascal}, ADE20K~\cite{zhou2019semantic}, MS COCO 2014~\cite{lin2014microsoft}, and Kvasir-SEG~\cite{jha2020kvasir} for segmentation.

\textbf{Implementation Details:}
The encoder is SigLIP~2 ViT-B/16~\cite{tschannen2025siglip2}, with $d=768$, $N_p=196$, $224\times224$ inputs, and $p=50$. We use $k\in\{0,\ldots,50\}$, $\mu=1$, and at most 200 ADMM iterations. Hyperparameters $\rho$ and $\gamma_s$ are selected from $\{0.01,0.05,0.1,0.2\}$.

\textbf{Evaluation Protocol:}
All CDFs, covariances, attention priors, eBIC choices, and precision matrices
are computed from the training partition only. Hyperparameters are chosen on
the validation partition and then frozen for test evaluation. Classification
uses accuracy and macro-F1 where reported; segmentation uses dataset-level
mean intersection-over-union. The controlled comparison uses the same input
resolution, frozen feature protocol, and data split for every baseline. In
contrast, task-specific numbers are taken under their published protocols and
are presented as contextual comparisons rather than controlled evidence.

\subsection{Controlled Comparison under Matched Protocol}
\label{sec:controlled}

Under the matched protocol (Table~\ref{tab:controlled}), CM-GLasso attains the
highest average classification accuracy (91.97\%) and improves the two-stage
graph baseline on both VOC and ADE20K. DINOv2 remains strongest on CIFAR-10
and Caltech-256, showing that the gain is not uniform across datasets.

\begin{strip}
\centering
\captionof{table}{Controlled comparison under frozen encoders, $224^2$ input, identical splits. ACC(\%) for classification and mIoU(\%) for segmentation.}
\label{tab:controlled}
\footnotesize
\setlength{\tabcolsep}{3pt}
\renewcommand{\arraystretch}{1.08}
\begin{tabular}{@{}llccccccl@{}}
\toprule
\textbf{Category} & \textbf{Method} & \textbf{C-10} & \textbf{C-100}
& \textbf{CUB} & \textbf{Cal.} & \textbf{Avg.} & \textbf{Interp.}
& \shortstack{\textbf{Seg.}\\\textbf{(VOC/ADE)}} \\
\midrule
\multirow{4}{*}{\shortstack{Frozen\\features}}
& DINOv2 lin. probe~\cite{oquab2024dinov2}
& \textbf{97.18} & 86.52 & 81.73 & \textbf{91.36} & 89.20 & low & ---/--- \\
& SigLIP~2 lin. probe
& 95.82 & 82.47 & 76.93 & 89.41 & 86.16 & low & 59.47/37.82 \\
& CoOp~\cite{zhou2022coop}
& 93.18 & 75.62 & 72.31 & 82.07 & 80.80 & low & ---/--- \\
& TIP-Adapter~\cite{zhang2022tipadapter}
& 93.65 & 76.48 & 73.86 & 83.15 & 81.79 & low & ---/--- \\
\midrule
\multirow{2}{*}{Graph}
& Std.\ GLasso+$\mathcal{H}_C$/$\mathcal{H}_S$
& 76.52 & 62.18 & 57.83 & 73.39 & 67.48 & high & 42.65/18.73 \\
& Two-Stage+$\mathcal{H}_C$/$\mathcal{H}_S$
& 92.47 & 91.83 & 87.26 & 85.48 & 89.26 & high & 68.17/51.46 \\
\midrule
\rowcolor{gray!12}
Ours & CM-GLasso+$\mathcal{H}_C$/$\mathcal{H}_S$
& 94.71 & \textbf{94.26} & \textbf{92.83} & 86.07
& \textbf{91.97} & high & \textbf{74.75}/\textbf{64.01} \\
\bottomrule
\end{tabular}
\end{strip}

\subsection{Comparison with Task-Specific Methods}
\label{sec:main_results}

Across the reported task-specific comparisons, CM-GLasso is strongest on the
listed classification and segmentation benchmarks
(Tables~\ref{tab:cls_results}--\ref{tab:seg_results}). Because these methods
use heterogeneous backbones and protocols, these tables complement rather
than replace the controlled comparison.

\begin{strip}
\centering
\begin{minipage}[t]{0.49\textwidth}
\centering
\captionof{table}{Comparison with representative task-specific classification methods.}
\label{tab:cls_results}
\footnotesize
\setlength{\tabcolsep}{2.5pt}
\renewcommand{\arraystretch}{1.05}
\begin{tabular}{@{}p{0.23\linewidth}p{0.43\linewidth}cc@{}}
\toprule
\textbf{Dataset} & \textbf{Method} & \textbf{F1} & \textbf{ACC}\\
\midrule
\multirow{5}{*}{\textbf{CUB-200}}
& ShuffleNetV2~\cite{ma2018shufflenetv2} & 0.8774 & 0.8763\\
& DA-VPT~\cite{ren2025davpt} & --- & 0.9020\\
& PRO-VPT~\cite{shang2025provpt} & --- & 0.9060\\
& VFPT~\cite{zeng2024vfpt} & --- & 0.9050\\
& \textbf{CM-GLasso (Ours)} & \textbf{0.8836} & \textbf{0.9283}\\
\midrule
\multirow{4}{*}{\textbf{CIFAR-100}}
& SSF~\cite{lian2022ssf} & --- & 0.9399\\
& Astroformer~\cite{dagli2023astroformer} & --- & 0.9360\\
& SPT-Swin~\cite{ferdous2024sptswin} & 0.9295 & 0.9295\\
& \textbf{CM-GLasso (Ours)} & \textbf{0.9300} & \textbf{0.9426}\\
\midrule
\multirow{4}{*}{\textbf{Caltech-256}}
& TMC~\cite{liu2023tmc} & --- & 0.8364\\
& CPC~\cite{zhi2024cpc} & --- & 0.8550\\
& EEG-VGG Fusion~\cite{jahanaray2025eeg} & --- & 0.8100\\
& \textbf{CM-GLasso (Ours)} & \textbf{0.8528} & \textbf{0.8607}\\
\bottomrule
\end{tabular}
\end{minipage}%
\hfill%
\begin{minipage}[t]{0.49\textwidth}
\centering
\captionof{table}{Comparative results on semantic segmentation (mIoU).}
\label{tab:seg_results}
\footnotesize
\setlength{\tabcolsep}{2.5pt}
\renewcommand{\arraystretch}{1.05}
\begin{tabular}{@{}p{0.28\linewidth}p{0.49\linewidth}c@{}}
\toprule
\textbf{Dataset} & \textbf{Method} & \textbf{mIoU}\\
\midrule
\multirow{4}{*}{\textbf{ADE20K}}
& OneFormer~\cite{jain2023oneformer} & 0.5700\\
& InternImage-H~\cite{wang2023internimage} & 0.6290\\
& OmniVec2~\cite{srivastava2024omnivec2} & 0.5850\\
& \textbf{CM-GLasso (Ours)} & \textbf{0.6401}\\
\midrule
\multirow{4}{*}{\textbf{Kvasir-SEG}}
& Polyp-PVT~\cite{dong2023polypvt} & 0.8640\\
& PolypMixNet~\cite{jia2024polypmixnet} & 0.8885\\
& MedFoundX~\cite{shawon2025medfoundx} & 0.8668\\
& \textbf{CM-GLasso (Ours)} & \textbf{0.8903}\\
\midrule
\multirow{4}{*}{\textbf{VOC-2012}}
& AuxSegNet+~\cite{xu2024auxsegnet} & 0.7090\\
& GroupViT~\cite{xu2022groupvit} & 0.5230\\
& PrivObNet~\cite{tay2024privobfnet} & 0.7150\\
& \textbf{CM-GLasso (Ours)} & \textbf{0.7475}\\
\midrule
\multirow{4}{*}{\textbf{COCO-2014}}
& MulP-VSS~\cite{duan2025mulpvss} & 0.4660\\
& CLIP-ES~\cite{lin2023clipes} & 0.4540\\
& BECO~\cite{rong2023beco} & 0.4510\\
& \textbf{CM-GLasso (Ours)} & \textbf{0.4682}\\
\bottomrule
\end{tabular}
\end{minipage}
\end{strip}

\subsection{Discussion of Main Results}

\textbf{Matched comparison.}
The controlled protocol is the most direct test of the proposed structural
estimator because it removes changes in resolution, split, and feature
training. CM-GLasso improves the two-stage graph baseline from 89.26\% to
91.97\% average classification accuracy. The largest controlled gains occur
in segmentation: $+6.58$ mIoU on VOC and $+12.55$ mIoU on ADE20K. Standard
class-wise GLasso is substantially worse, confirming that sparsity alone is
insufficient when class samples are limited. The comparison also identifies
an important boundary: DINOv2 remains better on CIFAR-10 and Caltech-256,
where a strong linear separator can be more effective than explicit graph
structure. The contribution is therefore not universal accuracy dominance,
but improved performance on structure-sensitive settings together with an
inspectable conditional-dependence representation.

\textbf{Task-specific classification.}
CM-GLasso reaches 0.9283 accuracy on CUB-200, compared with 0.9060 for the
strongest listed prompt-tuning competitor, and obtains 0.9426 on CIFAR-100.
The Caltech-256 margin is smaller (0.8607 versus 0.8550), which is consistent
with the controlled result showing that the dataset is already well served by
generic visual features. Macro-F1 follows the same trend where competing work
reports it. These numbers use the protocols of the cited papers, so they
indicate competitiveness rather than isolate the effect of CM-GLasso; the
matched table provides the causal comparison among model components.

\textbf{Task-specific segmentation.}
The method obtains 0.6401 mIoU on ADE20K, 0.8903 on Kvasir-SEG, 0.7475 on
VOC-2012, and 0.4682 on COCO-2014. Improvements are largest on VOC, whereas
Kvasir and COCO show narrow margins. This pattern suggests that the graph head
is most useful when long-range region consistency and class-conditioned
context complement local encoder evidence. The qualitative examples are
consistent with this interpretation: predictions preserve connected object
regions and suppress isolated background responses, although visualization
alone does not establish that every retained graph edge is semantically
meaningful.

\subsection{Ablation Studies}
\label{sec:ablation}

Table~\ref{tab:all_ablations} shows that shared-encoder rendering,
cross-attention nodes, Gaussianization, and joint optimization each matter;
combining common and specific precision components is also best. Performance
is stable near $\rho=.05,\gamma_s=.10$ (Table~\ref{tab:hyperparam}), while
eBIC frequently suppresses uninformative self-priors. Table~\ref{tab:complexity}
and Figures~\ref{fig:gam_vis}--\ref{fig:seg_vis} summarize cost and qualitative
behavior.

The ablations separate representation, distributional, and optimization
effects. Replacing rendered text and the shared SigLIP~2 pathway with
heterogeneous BERT/ViT encoders reduces ACC/mIoU from 91.97/68.65 to
84.23/53.27, while a CLIP-text/ViT pairing recovers only part of the gap.
Cross-attention nodes outperform both PCA and a linear fully connected map,
supporting the use of prototype-indexed footprints rather than dimensionality
reduction alone. Removing the nonparanormal transform lowers the held-out
normality pass rate and decreases both tasks, which empirically supports the
Gaussianization step without proving the full nonparanormal assumption.

Joint ADMM produces a larger common-structure ratio and a smaller
generalization gap than independent or two-stage estimation. Using only the
common component loses class-specific information, whereas using only
$\boldsymbol S^{(c)}$ discards reusable structure; their sum performs best.
Finally, text-to-image priors receive larger selected $k^*$ values and are less
frequently disabled than reverse or self-priors. This asymmetry supports text
as the auxiliary source, while the high $k^*=0$ ratio for self-priors verifies
that eBIC can revert to uniform penalization when a prior is unhelpful.

\begin{strip}
\centering
\captionof{table}{Comprehensive ablation studies. ACC and mIoU are reported in percent.}
\label{tab:all_ablations}
\footnotesize
\renewcommand{\arraystretch}{1.05}

\begin{minipage}[t]{0.485\textwidth}
\centering
\textbf{(a) Text Encoding Strategy}\par\vspace{1mm}
\begin{tabular*}{\linewidth}{@{\extracolsep{\fill}}lcc@{}}
\toprule
\textbf{Encoder} & \textbf{ACC} & \textbf{mIoU} \\
\midrule
BERT+ViT (Het.) & 84.23 & 53.27 \\
CLIP text+ViT~\cite{radford2021learning} & 88.02 & 59.89 \\
\rowcolor{gray!12}\textbf{Render+SigLIP 2} & \textbf{91.97} & \textbf{68.65} \\
\bottomrule
\end{tabular*}
\end{minipage}
\hfill
\begin{minipage}[t]{0.485\textwidth}
\centering
\textbf{(b) Patch-to-Node Mapping}\par\vspace{1mm}
\begin{tabular*}{\linewidth}{@{\extracolsep{\fill}}lcc@{}}
\toprule
\textbf{Strategy} & \textbf{ACC} & \textbf{mIoU} \\
\midrule
PCA ($768\!\to\!p$) & 70.86 & 44.18 \\
Linear FC & 87.37 & 62.93 \\
\rowcolor{gray!12}\textbf{Cross-Attn (Ours)} & \textbf{91.97} & \textbf{68.65} \\
\bottomrule
\end{tabular*}
\end{minipage}

\vspace{1mm}

\begin{minipage}[t]{0.485\textwidth}
\centering
\textbf{(c) Nonparanormal Transform}\par\vspace{1mm}
\begin{tabular*}{\linewidth}{@{\extracolsep{\fill}}lccc@{}}
\toprule
\textbf{Status} & \textbf{SW Pass} & \textbf{ACC} & \textbf{mIoU} \\
\midrule
w/o Trans. & $\sim23\%$ & 83.58 & 59.02 \\
\rowcolor{gray!12}\textbf{w/ Trans.} & \textbf{$\sim88\%$}
& \textbf{91.97} & \textbf{68.65} \\
\bottomrule
\end{tabular*}
\end{minipage}
\hfill
\begin{minipage}[t]{0.485\textwidth}
\centering
\textbf{(d) Optimization}\par\vspace{1mm}
\scriptsize
\setlength{\tabcolsep}{1.5pt}
\begin{tabular}{@{}p{0.67\linewidth}cc@{}}
\toprule
\textbf{Method} & \textbf{CSR} & \textbf{Gen. Gap} \\
\midrule
Indep. GLasso~\cite{friedman2008sparse} & --- & 8.29\% \\
Two-stage~\cite{danaher2014joint} & 0.37 & 3.04\% \\
\rowcolor{gray!12}\textbf{Joint ADMM} & \textbf{0.42} & \textbf{1.93\%} \\
\bottomrule
\end{tabular}
\end{minipage}

\vspace{1mm}

\begin{minipage}[t]{0.485\textwidth}
\centering
\textbf{(e) Task Head Precision Matrix}\par\vspace{1mm}
\begin{tabular*}{\linewidth}{@{\extracolsep{\fill}}lcc@{}}
\toprule
\textbf{Matrix Used} & \textbf{ACC} & \textbf{mIoU} \\
\midrule
Only $\boldsymbol{\Theta}_{\text{com}}$ & 84.82 & 63.17 \\
Only $\boldsymbol{S}^{(c)}$ & 88.43 & 65.58 \\
\rowcolor{gray!12}$\boldsymbol{\Theta}_{\text{com}}+\boldsymbol{S}^{(c)}$
& \textbf{91.97} & \textbf{68.65} \\
\bottomrule
\end{tabular*}
\end{minipage}
\hfill
\begin{minipage}[t]{0.485\textwidth}
\centering
\textbf{(f) Prior Selection via eBIC}\par\vspace{1mm}
\begin{tabular*}{\linewidth}{@{\extracolsep{\fill}}lcc@{}}
\toprule
\textbf{Direction} & $\bar{k}^{*}$ & \textbf{$k^*=0$ Ratio} \\
\midrule
Text $\to$ Image & 15.6 & 13.8\% \\
Image $\to$ Text & 6.7 & 33.1\% \\
\rowcolor{gray!12}\textbf{Self-Priors} & \textbf{$\approx0.1$} & \textbf{$>84.0\%$} \\
\bottomrule
\end{tabular*}
\end{minipage}

\vspace{2mm}

\begin{minipage}[t]{0.49\textwidth}
\centering
\caption{Sensitivity analysis of $\rho$ and $\gamma_s$. Values are ACC(\%)/mIoU(\%).}
\label{tab:hyperparam}
\scriptsize
\setlength{\tabcolsep}{2pt}
\renewcommand{\arraystretch}{1.05}
\resizebox{\linewidth}{!}{%
\begin{tabular}{@{}lcccc@{}}
\toprule
$\rho\downarrow\backslash\gamma_s\rightarrow$
&
0.01
&
0.05
&
0.10
&
0.20
\\
\midrule
0.01 & 90.53/66.94 & 91.02/67.53 & 91.18/67.72 & 90.61/67.03\\
0.05 & 91.27/67.68 & 91.64/68.27 & \textbf{91.97/68.65} & 91.41/67.91\\
0.10 & 91.12/67.56 & 91.53/68.14 & 91.82/68.47 & 91.23/67.63\\
0.20 & 90.46/66.81 & 90.83/67.24 & 91.04/67.40 & 90.21/66.59\\
\bottomrule
\end{tabular}
}
\end{minipage}%
\hfill%
\begin{minipage}[t]{0.49\textwidth}
\centering
\caption{Complexity and runtime on one NVIDIA A800.}
\label{tab:complexity}
\footnotesize
\setlength{\tabcolsep}{2.5pt}
\renewcommand{\arraystretch}{1.05}
\resizebox{\linewidth}{!}{%
\begin{tabular}{@{}lcc@{}}
\toprule
\textbf{Module} & \textbf{Time (sec)} & \textbf{Complexity} \\
\midrule
ViT Feature Extraction & 242 & $\mathcal{O}(NL(N_p^2d+N_pd^2))$ \\
Cross-Attention and Prior & 58 & $\mathcal{O}(NpN_pd_k+Cp^2N_p)$ \\
eBIC $k^*$ Selection & 268 & $\mathcal{O}(|\mathcal{K}|TCp^3)$ \\
Joint ADMM, fixed $k^*$ & 7.2 & $\mathcal{O}(TCp^3)$ \\
\midrule
Total Offline & $\sim575$ & --- \\
$\mathcal{H}_C$ Inference & --- & $\mathcal{O}(Cp^2)$ \\
$\mathcal{H}_S$ Inference & 6.8
& $\mathcal{O}(C(p^2d_k+pN_pd_k))$ \\
\bottomrule
\end{tabular}
}
\end{minipage}

\vspace{2mm}


  \begin{minipage}[t]{0.49\textwidth}
  \centering

  \newlength{\clsw}
  \setlength{\clsw}{0.108\linewidth}

  \setlength{\tabcolsep}{1.2pt}
  \renewcommand{\arraystretch}{0.6}

  \begin{tabular}{
    @{}
    r
    c@{\hskip 1.2pt}
    c@{\hskip 1.2pt}
    c@{\hskip 1.2pt}
    c
    @{\hskip 5pt}
    c@{\hskip 1.2pt}
    c@{\hskip 1.2pt}
    c@{\hskip 1.2pt}
    c
    @{}
  }

    &
    \multicolumn{4}{c}{\scriptsize\textbf{Input Images}}
    &
    \multicolumn{4}{c}{\scriptsize\textbf{GAM Visualizations}}
    \\[3pt]

    \raisebox{0.45\clsw}[0pt][0pt]{%
      \rotatebox[origin=c]{90}{\scriptsize\textbf{CUB-200}}%
    }
    &
    \safeincludegraphics[width=\clsw,height=\clsw]{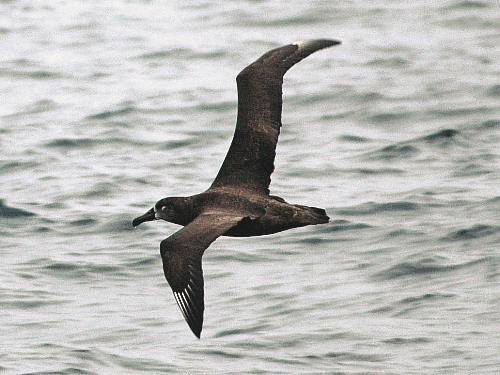}
    &
    \safeincludegraphics[width=\clsw,height=\clsw]{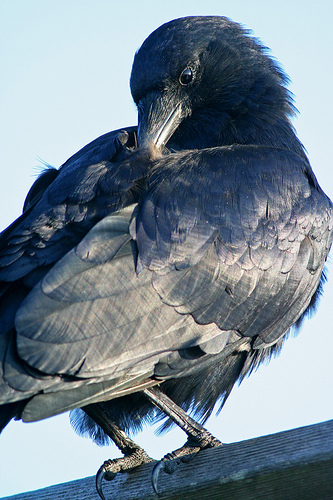}
    &
    \safeincludegraphics[width=\clsw,height=\clsw]{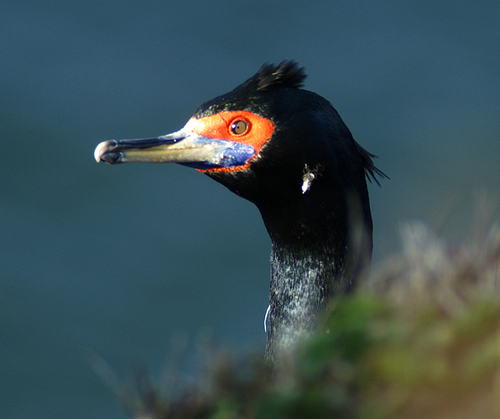}
    &
    \safeincludegraphics[width=\clsw,height=\clsw]{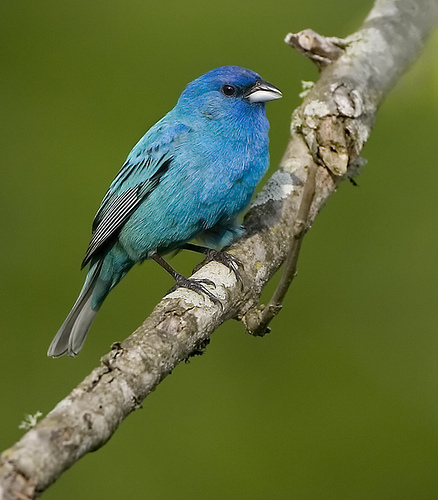}
    &
    \safeincludegraphics[width=\clsw,height=\clsw]{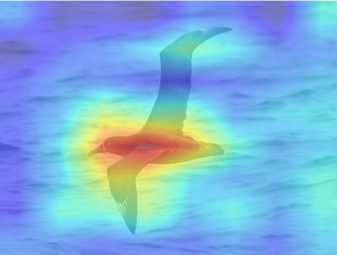}
    &
    \safeincludegraphics[width=\clsw,height=\clsw]{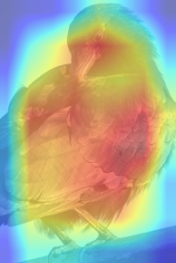}
    &
    \safeincludegraphics[width=\clsw,height=\clsw]{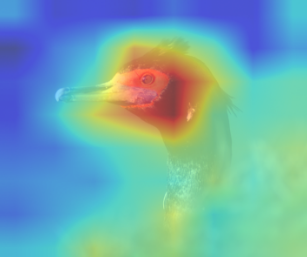}
    &
    \safeincludegraphics[width=\clsw,height=\clsw]{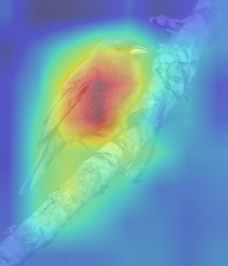}
    \\[2pt]

    \raisebox{0.45\clsw}[0pt][0pt]{%
      \rotatebox[origin=c]{90}{\scriptsize\textbf{CIFAR-10}}%
    }
    &
    \safeincludegraphics[width=\clsw,height=\clsw]{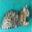}
    &
    \safeincludegraphics[width=\clsw,height=\clsw]{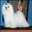}
    &
    \safeincludegraphics[width=\clsw,height=\clsw]{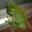}
    &
    \safeincludegraphics[width=\clsw,height=\clsw]{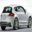}
    &
    \safeincludegraphics[width=\clsw,height=\clsw]{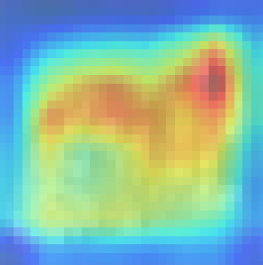}
    &
    \safeincludegraphics[width=\clsw,height=\clsw]{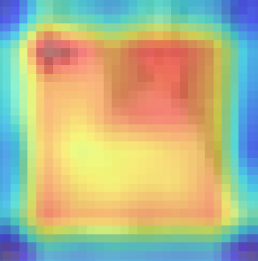}
    &
    \safeincludegraphics[width=\clsw,height=\clsw]{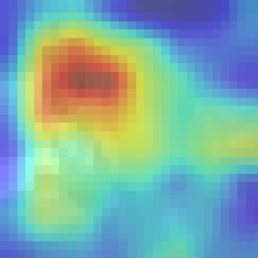}
    &
    \safeincludegraphics[width=\clsw,height=\clsw]{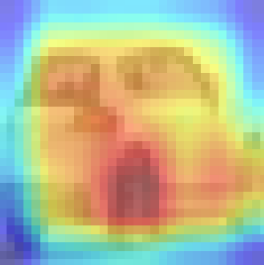}
    \\[2pt]

    \raisebox{0.45\clsw}[0pt][0pt]{%
      \rotatebox[origin=c]{90}{\scriptsize\textbf{CIFAR-100}}%
    }
    &
    \safeincludegraphics[width=\clsw,height=\clsw]{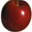}
    &
    \safeincludegraphics[width=\clsw,height=\clsw]{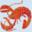}
    &
    \safeincludegraphics[width=\clsw,height=\clsw]{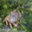}
    &
    \safeincludegraphics[width=\clsw,height=\clsw]{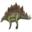}
    &
    \safeincludegraphics[width=\clsw,height=\clsw]{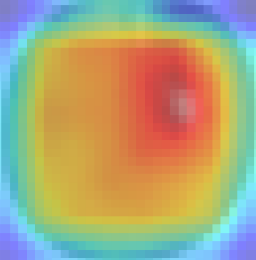}
    &
    \safeincludegraphics[width=\clsw,height=\clsw]{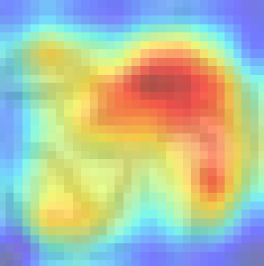}
    &
    \safeincludegraphics[width=\clsw,height=\clsw]{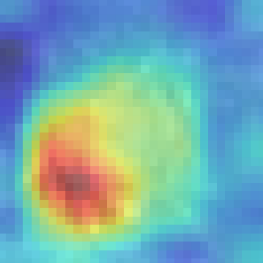}
    &
    \safeincludegraphics[width=\clsw,height=\clsw]{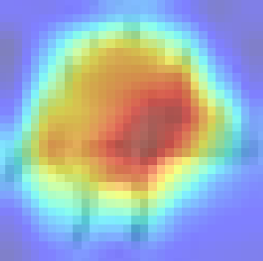}
    \\[2pt]

    \raisebox{0.45\clsw}[0pt][0pt]{%
      \rotatebox[origin=c]{90}{\scriptsize\textbf{Caltech-256}}%
    }
    &
    \safeincludegraphics[width=\clsw,height=\clsw]{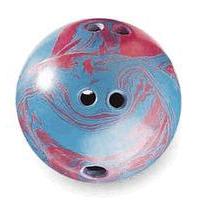}
    &
    \safeincludegraphics[width=\clsw,height=\clsw]{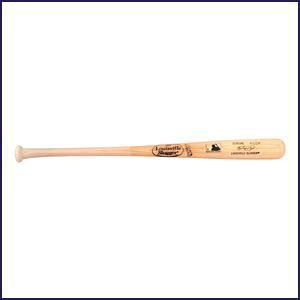}
    &
    \safeincludegraphics[width=\clsw,height=\clsw]{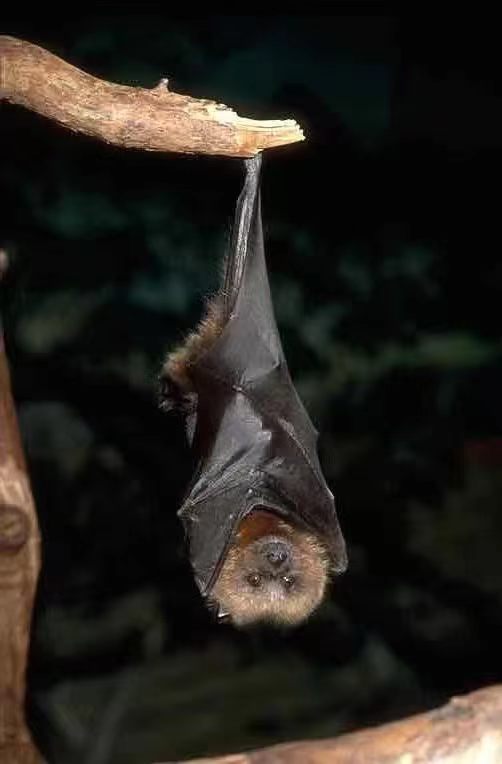}
    &
    \safeincludegraphics[width=\clsw,height=\clsw]{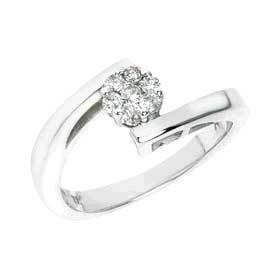}
    &
    \safeincludegraphics[width=\clsw,height=\clsw]{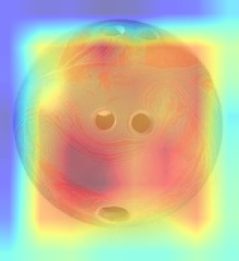}
    &
    \safeincludegraphics[width=\clsw,height=\clsw]{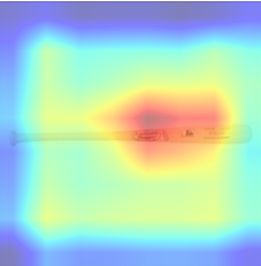}
    &
    \safeincludegraphics[width=\clsw,height=\clsw]{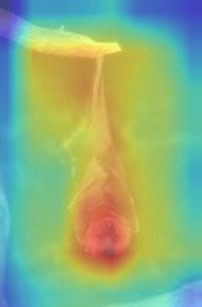}
    &
    \safeincludegraphics[width=\clsw,height=\clsw]{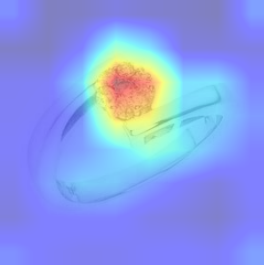}
    \\

  \end{tabular}

  \vspace{0.5mm}

  \captionof{figure}{%
    \textbf{GAM visualization of the classification head $\mathcal{H}_C$.}
    Across four datasets, warm regions emphasize class-discriminative
    objects and contours rather than background clutter.%
  }
  \label{fig:gam_vis}
  \end{minipage}%
  \hfill%
  \begin{minipage}[t]{0.49\textwidth}
  \centering

  \newlength{\segw}
  \setlength{\segw}{0.095\linewidth}

  \setlength{\tabcolsep}{0.8pt}
  \renewcommand{\arraystretch}{0.5}

  \begin{tabular}{
    @{}
    r
    c@{\hskip 0.8pt}
    c@{\hskip 0.8pt}
    c
    @{\hskip 4pt}
    c@{\hskip 0.8pt}
    c@{\hskip 0.8pt}
    c
    @{\hskip 4pt}
    c@{\hskip 0.8pt}
    c@{\hskip 0.8pt}
    c
    @{}
  }

    &
    \multicolumn{3}{c}{\scriptsize\textbf{Sample 1}}
    &
    \multicolumn{3}{c}{\scriptsize\textbf{Sample 2}}
    &
    \multicolumn{3}{c}{\scriptsize\textbf{Sample 3}}
    \\[-0.5pt]

    &
    {\tiny Image} & {\tiny GT} & {\tiny Ours}
    &
    {\tiny Image} & {\tiny GT} & {\tiny Ours}
    &
    {\tiny Image} & {\tiny GT} & {\tiny Ours}
    \\[2pt]

    \raisebox{0.45\segw}[0pt][0pt]{%
      \rotatebox[origin=c]{90}{\scriptsize\textbf{ADE20K}}%
    }
    &
    \safeincludegraphics[width=\segw,height=\segw]{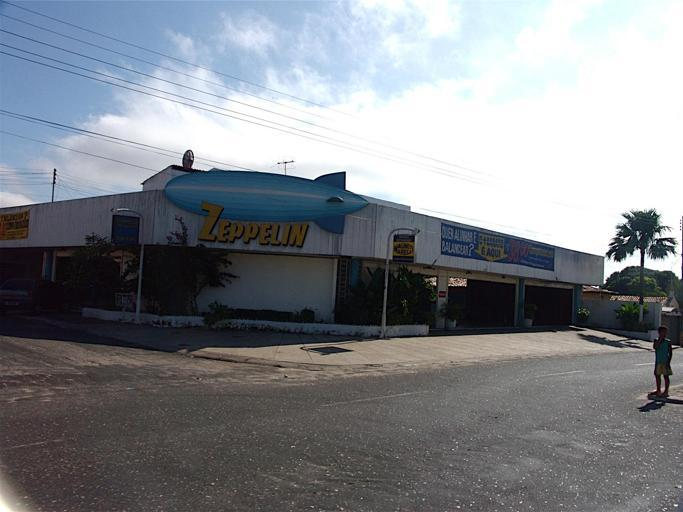}
    &
    \safeincludegraphics[width=\segw,height=\segw]{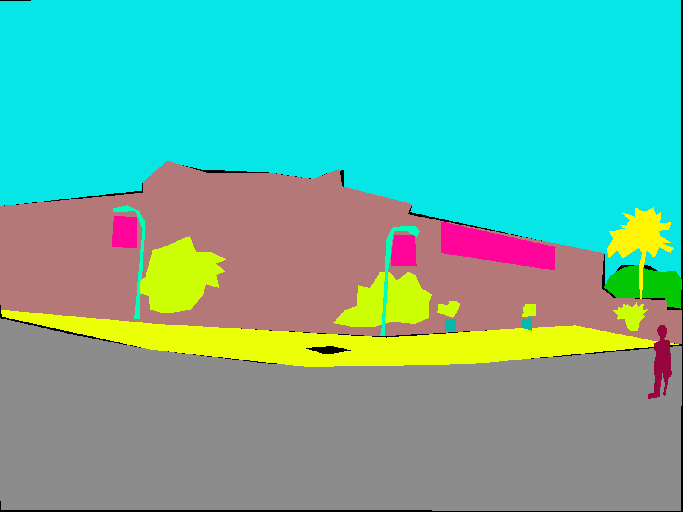}
    &
    \safeincludegraphics[width=\segw,height=\segw]{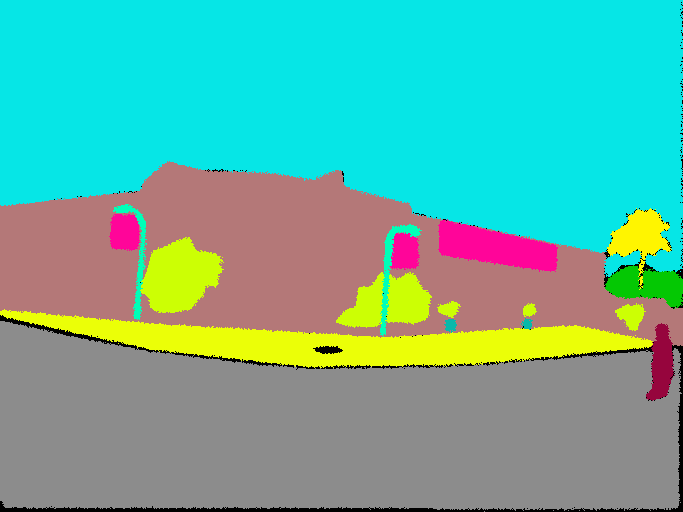}
    &
    \safeincludegraphics[width=\segw,height=\segw]{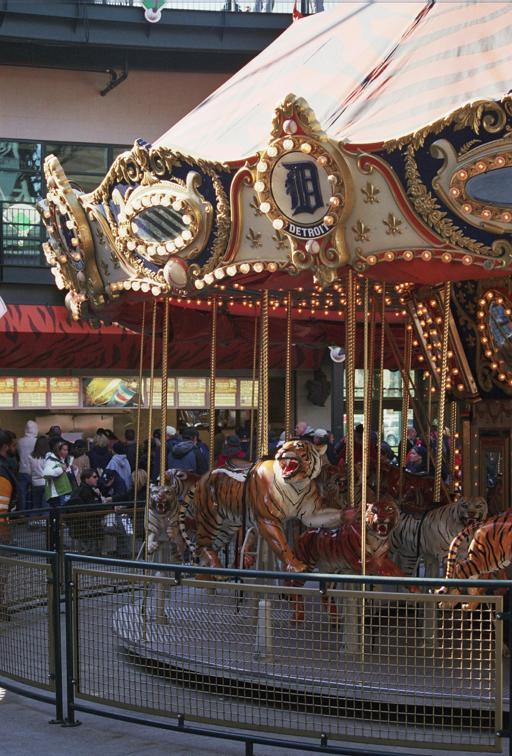}
    &
    \safeincludegraphics[width=\segw,height=\segw]{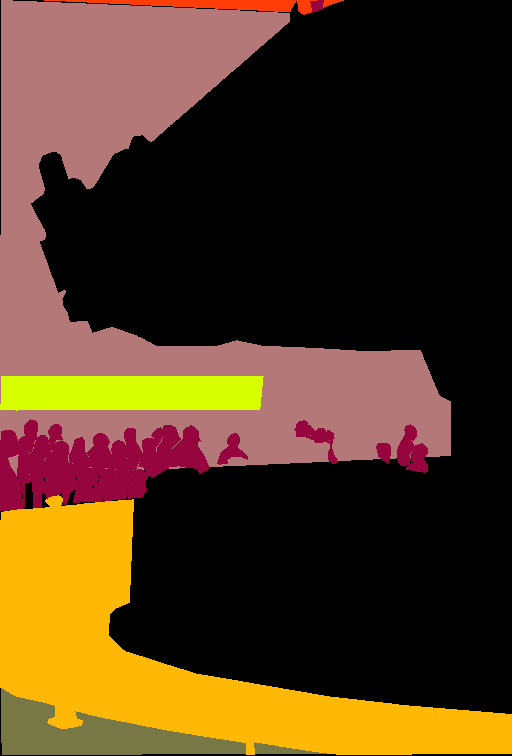}
    &
    \safeincludegraphics[width=\segw,height=\segw]{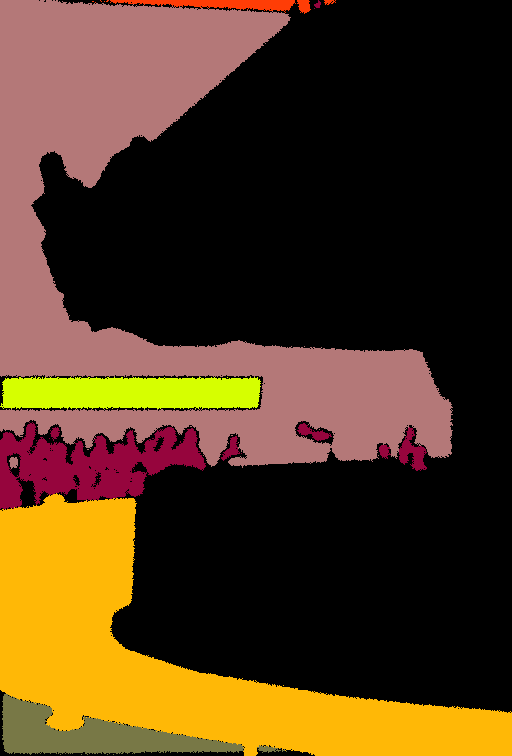}
    &
    \safeincludegraphics[width=\segw,height=\segw]{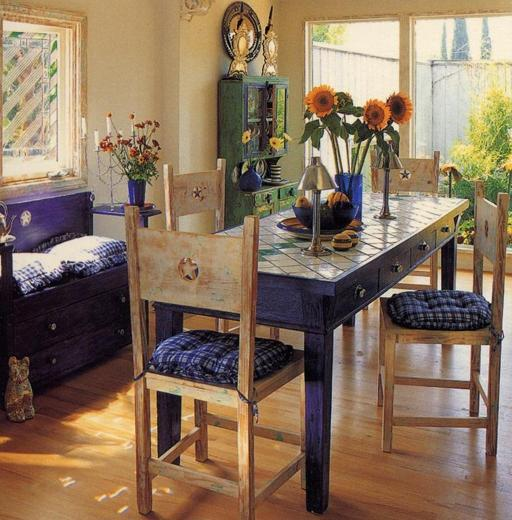}
    &
    \safeincludegraphics[width=\segw,height=\segw]{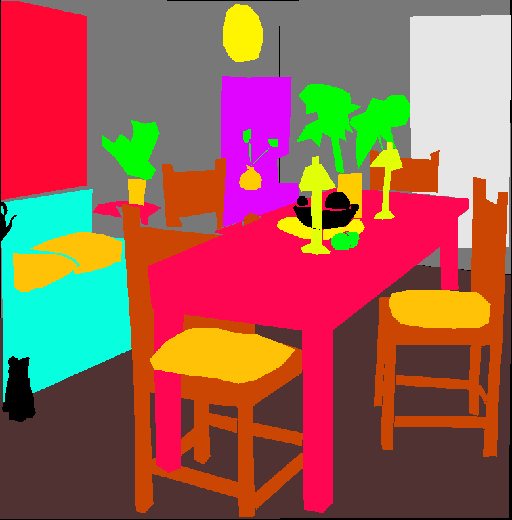}
    &
    \safeincludegraphics[width=\segw,height=\segw]{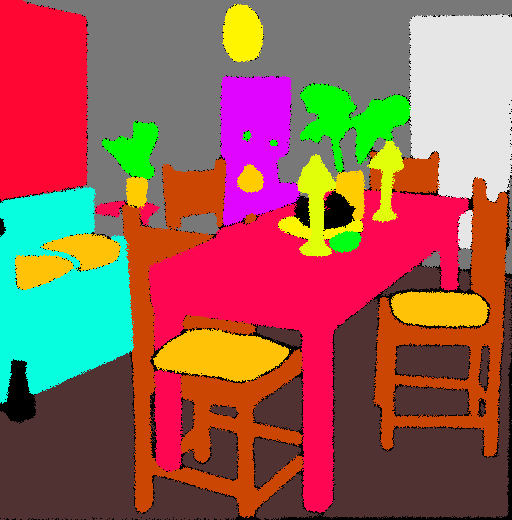}
    \\[2pt]

    \raisebox{0.45\segw}[0pt][0pt]{%
      \rotatebox[origin=c]{90}{\scriptsize\textbf{Kvasir}}%
    }
    &
    \safeincludegraphics[width=\segw,height=\segw]{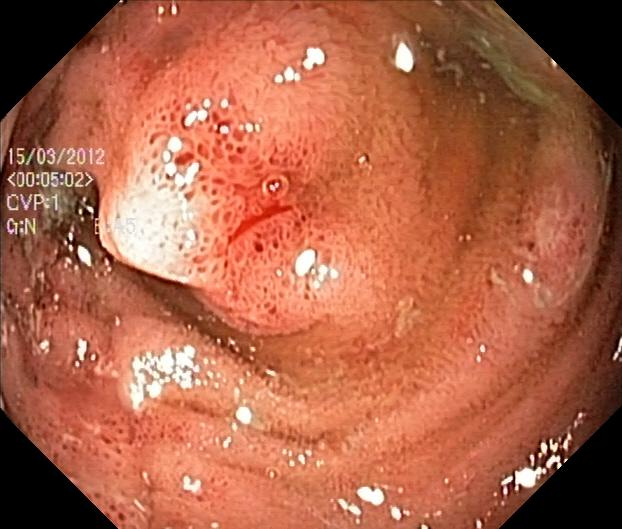}
    &
    \safeincludegraphics[width=\segw,height=\segw]{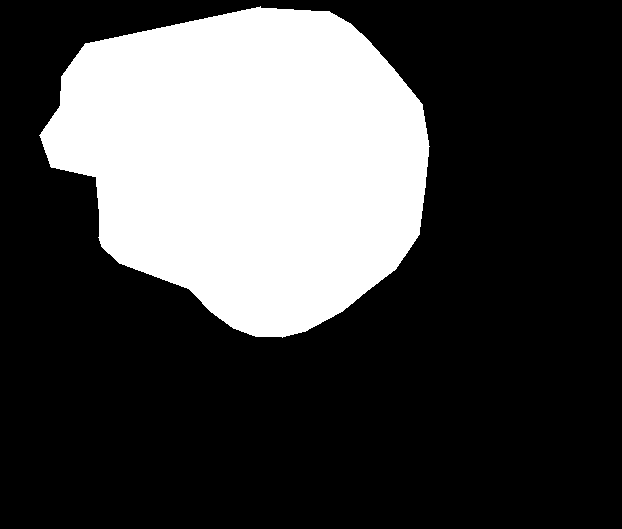}
    &
    \safeincludegraphics[width=\segw,height=\segw]{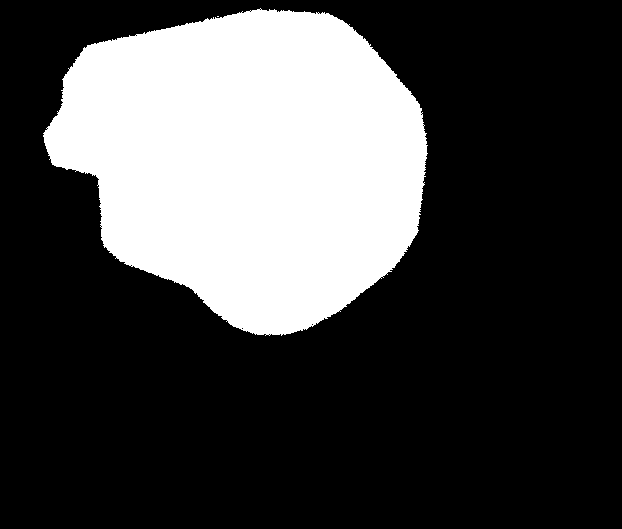}
    &
    \safeincludegraphics[width=\segw,height=\segw]{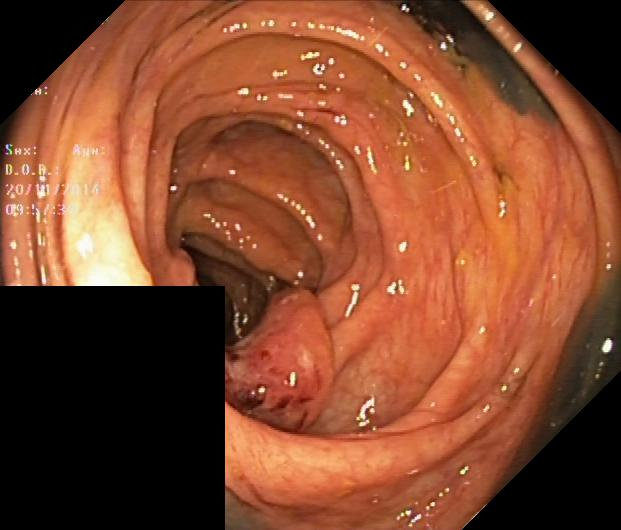}
    &
    \safeincludegraphics[width=\segw,height=\segw]{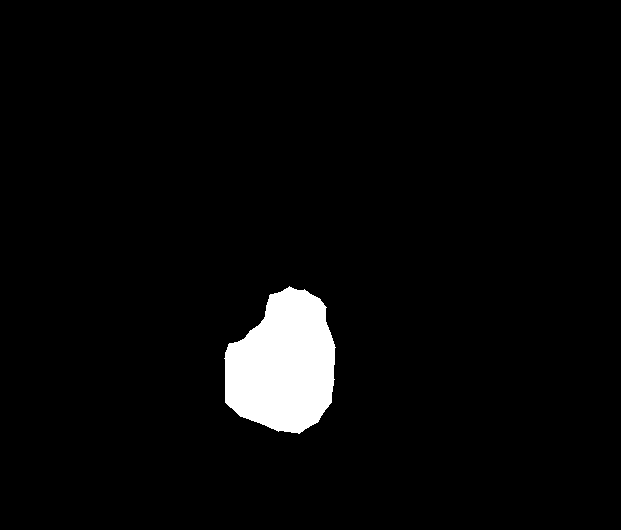}
    &
    \safeincludegraphics[width=\segw,height=\segw]{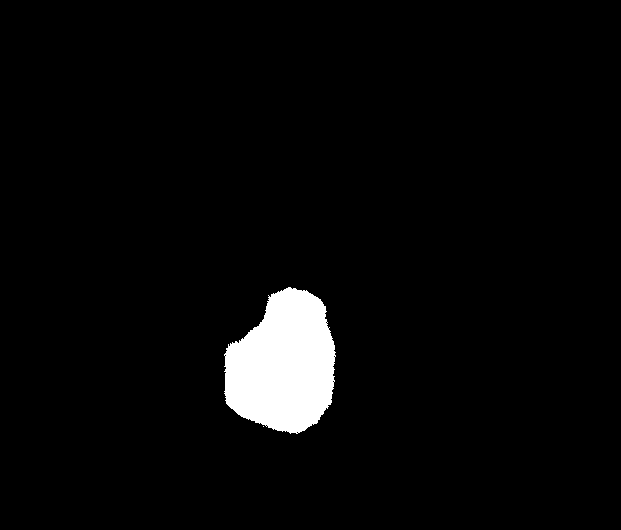}
    &
    \safeincludegraphics[width=\segw,height=\segw]{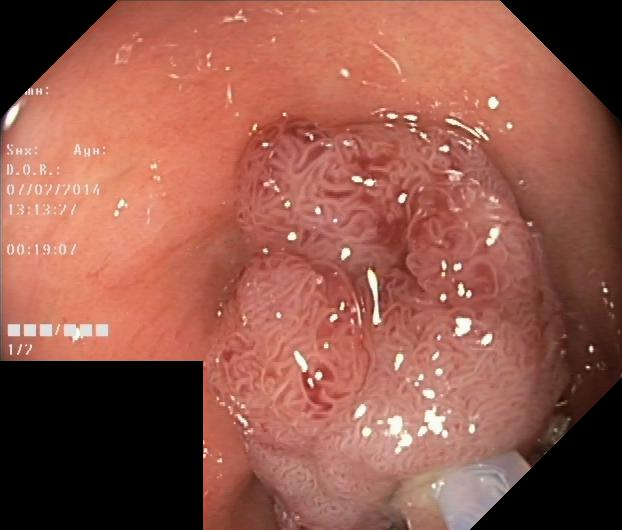}
    &
    \safeincludegraphics[width=\segw,height=\segw]{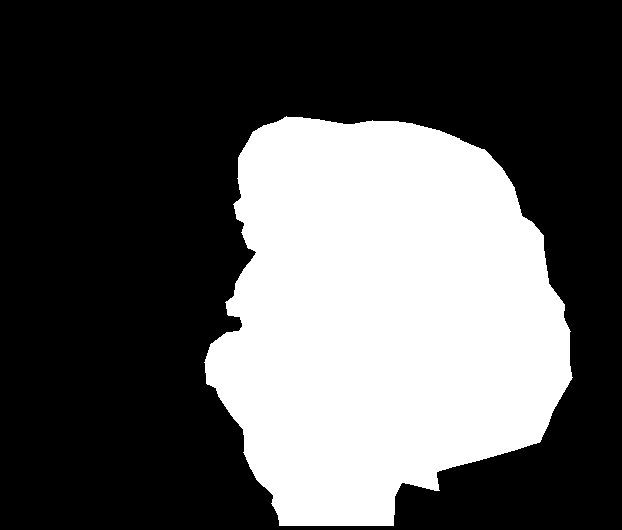}
    &
    \safeincludegraphics[width=\segw,height=\segw]{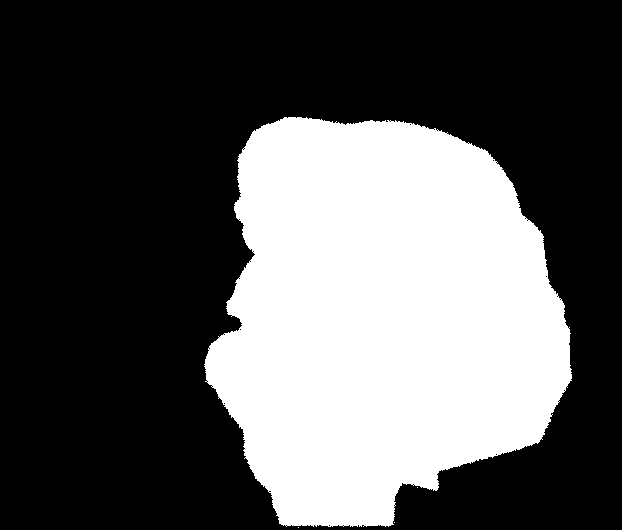}
    \\[2pt]

    \raisebox{0.45\segw}[0pt][0pt]{%
      \rotatebox[origin=c]{90}{\scriptsize\textbf{VOC-2012}}%
    }
    &
    \safeincludegraphics[width=\segw,height=\segw]{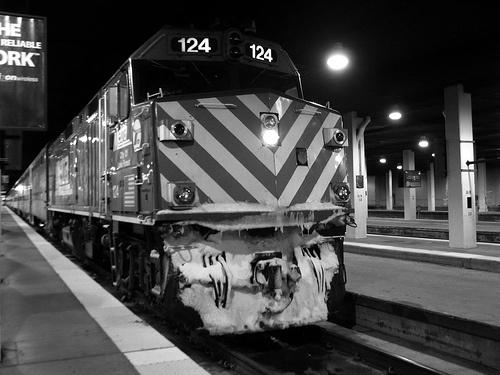}
    &
    \safeincludegraphics[width=\segw,height=\segw]{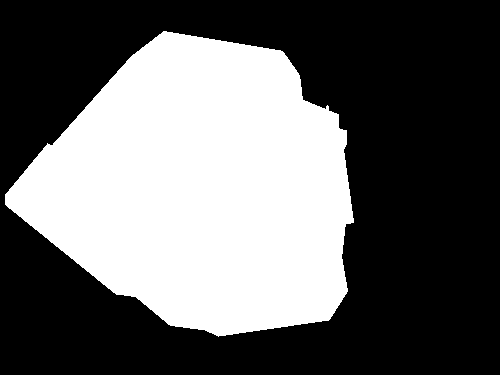}
    &
    \safeincludegraphics[width=\segw,height=\segw]{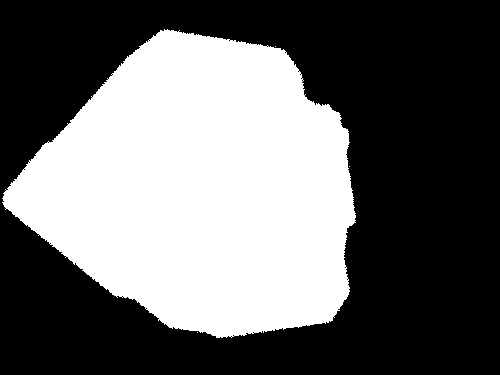}
    &
    \safeincludegraphics[width=\segw,height=\segw]{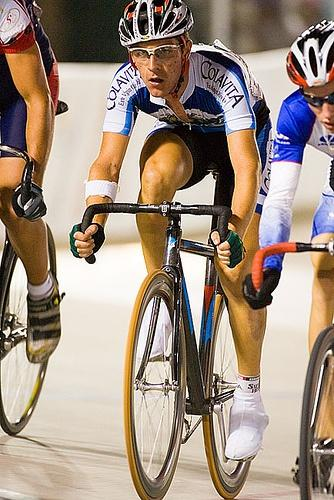}
    &
    \safeincludegraphics[width=\segw,height=\segw]{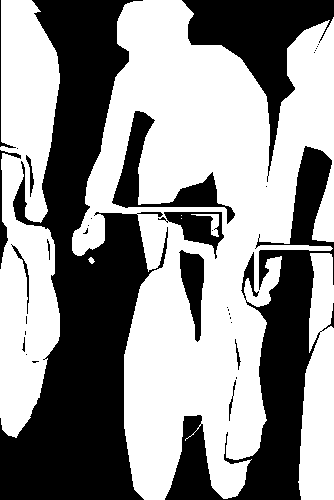}
    &
    \safeincludegraphics[width=\segw,height=\segw]{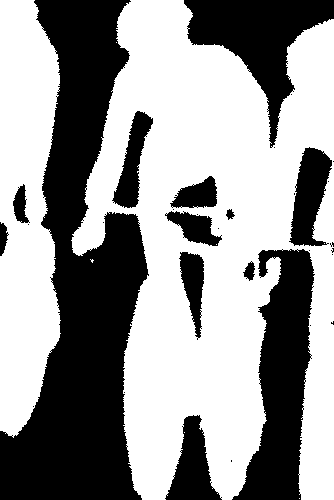}
    &
    \safeincludegraphics[width=\segw,height=\segw]{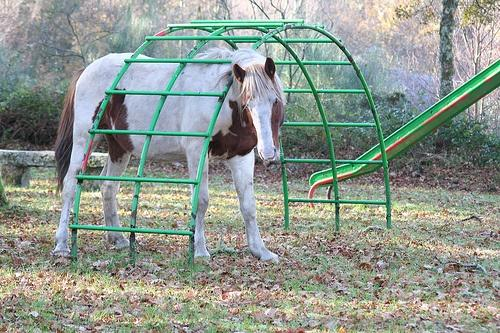}
    &
    \safeincludegraphics[width=\segw,height=\segw]{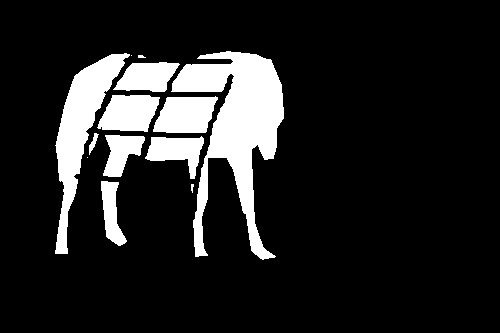}
    &
    \safeincludegraphics[width=\segw,height=\segw]{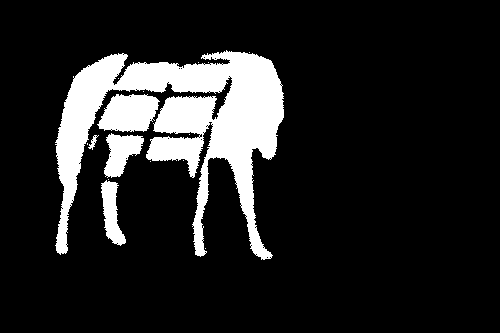}
    \\[2pt]

    \raisebox{0.45\segw}[0pt][0pt]{%
      \rotatebox[origin=c]{90}{\scriptsize\textbf{COCO}}%
    }
    &
    \safeincludegraphics[width=\segw,height=\segw]{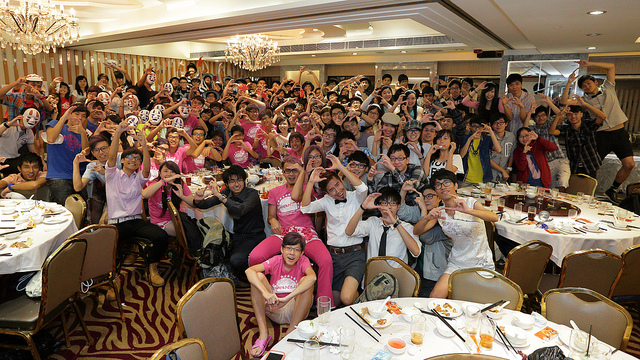}
    &
    \safeincludegraphics[width=\segw,height=\segw]{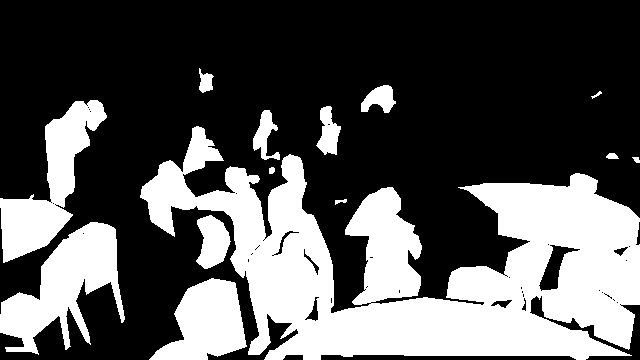}
    &
    \safeincludegraphics[width=\segw,height=\segw]{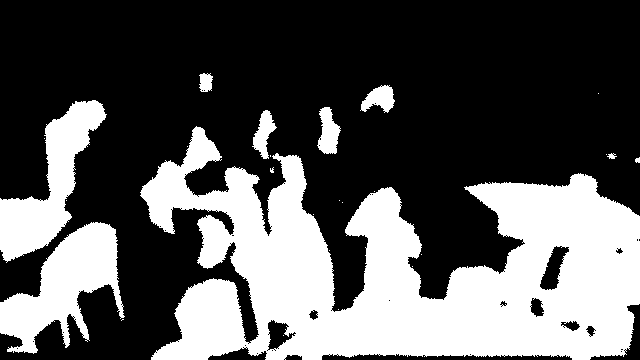}
    &
    \safeincludegraphics[width=\segw,height=\segw]{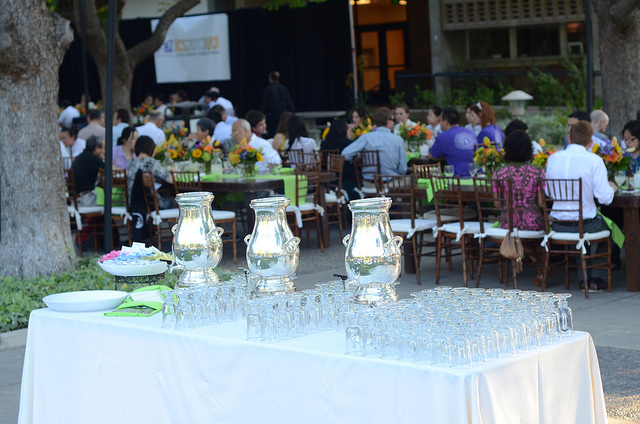}
    &
    \safeincludegraphics[width=\segw,height=\segw]{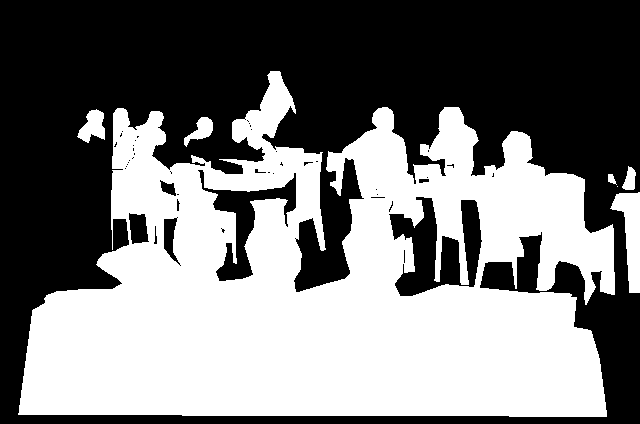}
    &
    \safeincludegraphics[width=\segw,height=\segw]{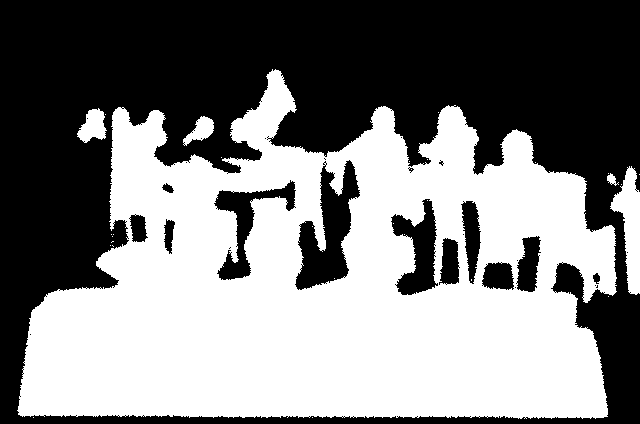}
    &
    \safeincludegraphics[width=\segw,height=\segw]{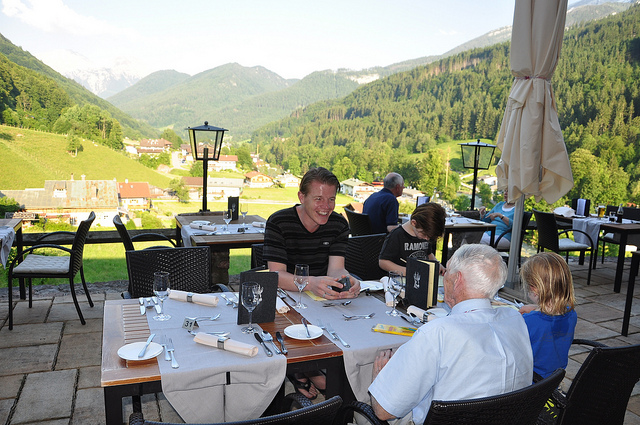}
    &
    \safeincludegraphics[width=\segw,height=\segw]{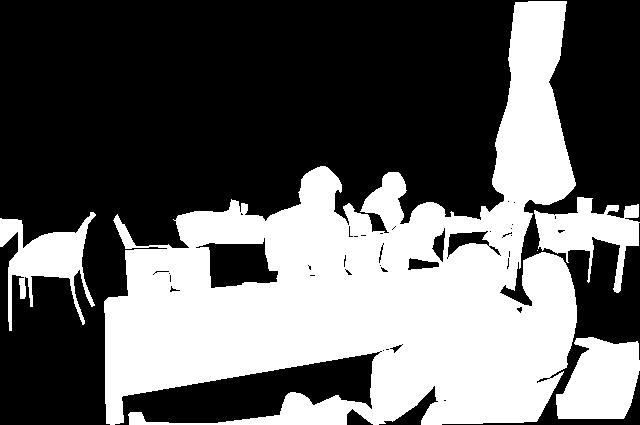}
    &
    \safeincludegraphics[width=\segw,height=\segw]{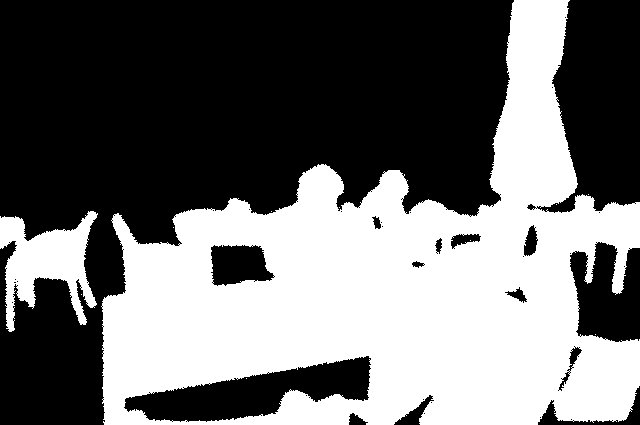}
    \\

  \end{tabular}

  \vspace{0.5mm}

  \captionof{figure}{%
    \textbf{Qualitative segmentation results of $\mathcal{H}_S$.}
    Image/GT/prediction triplets show coherent regions and boundaries on
    ADE20K, Kvasir-SEG, VOC-2012, and COCO-2014.%
  }
  \label{fig:seg_vis}
  \end{minipage}

  \vspace{0.5mm}

  \begin{minipage}[t]{0.49\textwidth}
  \refstepcounter{section}
  {\centering\large\bfseries \thesection\quad Conclusion\par}
  \vspace{0.5mm}
  CM-GLasso uses shared prototype attention to construct cross-modal priors
  and jointly estimates common and class-specific sparse precision matrices.
  Across eight benchmarks, it remains competitive while exposing interpretable
  graph topology. The controlled results show that its advantage is strongest
  on structure-sensitive tasks rather than uniform across all datasets. This
  distinction is important: the method is intended to complement strong
  foundation-model features with an explicit conditional-dependence layer,
  not to replace feature learning. The learned decomposition also supports two
  views of the data---a reusable common graph and class-dependent deviations---
  while the eBIC mechanism can suppress an auxiliary prior when it is not
  supported by training observations. This combination makes the learned
  structure useful not only for prediction, but also for inspecting how
  cross-modal evidence changes class-specific dependencies.
  \end{minipage}%
  \hfill%
  \begin{minipage}[t]{0.49\textwidth}
  \refstepcounter{subsection}
  {\centering\large\bfseries \thesubsection\quad Limitations\par}
  \vspace{0.5mm}
  ADMM requires eigendecompositions and its three-block form lacks a general
  convergence guarantee. Rendering is typography-dependent, the discriminant
  omits a Jacobian, diagonal loading is needed for $n_c<p$, and training is
  decoupled. In addition, prototype nodes are learned for predictive utility
  and therefore need not correspond one-to-one with human concepts. Attention
  overlap is only a heuristic structural prior and cannot be interpreted as a
  causal or semantic edge. The task-specific comparisons also combine results
  from heterogeneous published protocols; only the matched comparison isolates
  the effect of the graph estimator. Finally, the current experiments use a
  fixed graph dimension and do not characterize scaling to very large class
  vocabularies or substantially denser patch grids.
  \end{minipage}

  \vspace{2mm}

  \begin{minipage}[t]{0.49\textwidth}
  \refstepcounter{subsection}
  {\centering\large\bfseries \thesubsection\quad Practical Implications\par}
  \vspace{0.5mm}
  The framework is most suitable when an application values an inspectable
  interaction structure in addition to predictive performance. Graph fitting
  is offline, and Table~\ref{tab:complexity} shows that feature extraction and
  model selection dominate its cost; the fixed-$k^*$ ADMM stage is relatively
  small. Once estimated, the same precision matrices can be reused by both
  classification and segmentation heads without refitting the encoder. The
  common--specific decomposition also provides a direct diagnostic: common
  edges summarize recurring dependencies, whereas large class-specific
  deviations indicate where a category departs from the shared organization.
  Such diagnostics should be validated against domain knowledge before being
  used for scientific conclusions.
  \end{minipage}%
  \hfill%
  \begin{minipage}[t]{0.49\textwidth}
  \refstepcounter{subsection}
  {\centering\large\bfseries \thesubsection\quad Future Work\par}
  \vspace{0.5mm}
  Future work will study a provably convergent two-block reformulation and
  implicit differentiation through the graph solver for end-to-end training.
  A Jacobian-aware nonparanormal discriminant would make class comparisons
  statistically better founded, while typography augmentation or direct
  rendered-text pretraining could improve robustness of the text pathway.
  Adaptive selection of the graph dimension may reduce eigendecomposition cost
  and avoid fixing $p$ across datasets. Further priorities include uncertainty
  estimates for individual edges, stability selection across data resamples,
  calibration of graph explanations, and evaluation on additional modalities
  where auxiliary attention footprints may provide complementary structural
  evidence.
  \end{minipage}
\end{strip}

\clearpage
\raggedbottom

\bibliography{aaai2027}

\end{document}